\documentclass[conference]{IEEEtran}
\IEEEoverridecommandlockouts
\usepackage{amsmath}
\usepackage{times}
\usepackage{graphicx}
\usepackage{multirow}
\usepackage[none]{hyphenat}
\usepackage{float}
\usepackage[caption=false]{subfig}
\usepackage{acro}
\usepackage{siunitx}
\usepackage{hyperref}
\usepackage[nameinlink,capitalise]{cleveref}
\usepackage{tikz}
\usepackage{textcomp}
\DeclareAcronym{fmcw}{short=FMCW, long=frequency modulated continuous wave}
\DeclareAcronym{fft}{short=FFT, long=fast Fourier transform}
\DeclareAcronym{rnn}{short=RNN, long=recurrent neural network}
\DeclareAcronym{gru}{short=GRU, long=gated recurrent unit}
\DeclareAcronym{rdi}{short=RDI, long=range-Doppler image}
\newcommand\copyrighttext{%
    \footnotesize \textcopyright 2024 IEEE. Personal use of this material is permitted. Permission from IEEE must be
    obtained for all other uses, in any current or future media, including
    reprinting/republishing this material for advertising or promotional purposes, creating new
    collective works, for resale or redistribution to servers or lists, or reuse of any copyrighted
    component of this work in other works.}
\newcommand\copyrightnotice{%
\begin{tikzpicture}[remember picture,overlay]
\node[anchor=south,yshift=10pt] at (current page.south) {\fbox{\parbox{\dimexpr\textwidth-\fboxsep-\fboxrule\relax}{\copyrighttext}}};
\end{tikzpicture}%
}
\begin{document}
\raggedbottom
%
%
%
\title{Gesture Recognition for FMCW Radar on the Edge}
%
%
\author{%
    \IEEEauthorblockN{%
        Maximilian Strobel*,
        Stephan Schoenfeldt*,
        Jonas Daugalas*
    }
    \IEEEauthorblockA{%
        *Infineon Technologies AG, Munich, Germany\\
        Maximilian.Strobel@infineon.com\\
    }
}
%
\maketitle
\copyrightnotice
%
%
\begin{abstract}
    This paper introduces a lightweight gesture recognition system based on 60 GHz \ac{fmcw} radar.
    We show that gestures can be characterized efficiently by a set of five features, and propose a slim radar processing algorithm to extract these features.
    In contrast to previous approaches, we avoid heavy 2D processing, i.e. range-Doppler imaging, and perform instead an early target detection - this allows us to port the system to fully embedded platforms with tight constraints on memory, compute and power consumption.
    A \ac{rnn} based architecture exploits these features to jointly detect and classify five different gestures.
    The proposed system recognizes gestures with an F1 score of 98.4\% on our hold-out test dataset, it runs on an Arm® Cortex®-M4 microcontroller requiring less than 280 kB of flash memory, 120 kB of RAM, and consuming 75 mW of power.
\end{abstract}
%
%

\section{Introduction}
Gesture recognition systems provide an uncomplicated and intuitive modality for human machine interfaces.
Unlike physical buttons, switches and touch screens, gesture-controlled systems are touch-less, which improves their hygiene.
The complexity of gestures limits those systems to simple control tasks compared to the more feature-rich touch-screens and voice assistants.
On the other hand, this simplicity allows a less cognitive demanding control, while the user can focus on other things.
This makes gesture sensing an ideal candidate for controlling various consumer and IoT devices like intelligent thermostats \cite{Google:Nest}, smart TVs or other smart home appliances \cite{Infineon:Soli}.

In this work, we focus on 60 GHz \ac{fmcw} based radar sensing, specifically on Soli - a miniature radar device that was co-developed by Infineon and Google \cite{Trotta:Soli}.
The small form-factor of the Soli sensor allows an integration into IoT devices with space limitations.
Additionally, the sensor is tuned for efficiency and low power consumption \cite{Trotta:Soli}, which is essential for IoT and consumer applications.
While image sensors offer a higher resolution to resolve gestures, these systems suffer from privacy issues, high cost, high power consumption, and higher compute and memory requirements.
We show that the reduced information in the radar signal is sufficient for robust gesture recognition, while still benefiting from the advantages mentioned earlier.

Early studies of the Soli sensor propose a feature extraction based on \acp{rdi} followed by a random forest classifier \cite{Google:Soli}.
Other authors propose for 24 GHz \ac{fmcw} radar a similar signal processing combined with an \ac{rnn} to encode temporal characteristics of gestures \cite{Choi:LSTM}.
The majority of work replaces those hand-crafted processing chains with neural networks, which jointly extract features and classify gestures \cite{Dekker:CNN}.
In RadarNet \cite{Google:RadarNet} a convolutional neural network, called Frame Model, extracts features on a frame basis and aggregates the information using an \ac{rnn}, namely Temporal Model.
All those approaches rely on heavy 2D data processing, either as preparation for feature extraction \cite{Google:Soli,Choi:LSTM} or as input to neural networks \cite{Dekker:CNN,Google:RadarNet}.

Our main contributions are:
(1) a lightweight radar processing algorithm, which requires significantly less computational resources than conventional approaches;
(2) a tiny neural network enhanced with a custom label refinement and data augmentation strategy to detect and classify robustly five different gestures.

Finally, we conclude the results of our work and give some directions for future work.

\section{Radar Processing}
\begin{figure}[!htbp]
    \centering
    \includegraphics[width=\columnwidth]{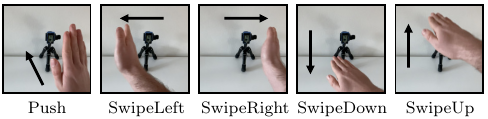}
    \caption{
        The set of gestures used in this work.
    }
    \label{fig:gestures}
\end{figure}

We examine in this work gestures done by persons close to the radar device.
The subjects of this study perform the gestures towards the sensor in a field of view of $\pm\SI{30}{\degree}$ at a distance of one meter or less.
The set of gestures contains four directional swipes and a push gesture towards the sensor, see \cref{fig:gestures}.
We claim that those gestures can be fully described by a time series of RF scattering characteristics of the moving hand:
\begin{itemize}
    \item   radial distance
    \item	radial velocity
    \item	horizontal angle
    \item	vertical angle
    \item   signal magnitude
\end{itemize}
Based on this assumption, the proposed radar processing algorithm identifies the hand as a target first, followed by an extraction of those characteristics.

We configure the radar device to emit chirps spanning a frequency range from $f_{low} = \SI{58.5}{\giga \hertz}$ to $f_{high} = \SI{62.5}{\giga \hertz}$ inducing a range resolution $\Delta r = \SI{37.5}{\milli \meter}$.
The device receives the reflected RF signal on three receive antennas, arranged in an L-shape, which allow to estimate the angle of the scattering target in two planes.
The received signal of every chirp is converted to an intermediate frequency, anti-aliasing filtered, and digitized with 64 samples at $\SI{2}{\mega \hertz}$ leading to a maximum resolvable range $r_{max} = \SI{1.2}{\meter}$.
The system sends bursts of chirps with a frame rate of $\SI{33.3}{\hertz}$ wherein the individual chirps are separated by a pulse repetition time $T_{PRT}$ of $\SI{300}{\micro\second}$.
The bursts contain 32 chirps, which results in a 3D array [R x C x S] with $R=3$ receive channels, the \textit{slow time} axis $C=32$, and \textit{fast time} axis $S=64$.

\subsection{Target detection}
In order to detect the hand as a moving target, we start to transform the raw radar via fast time processing into \textit{range profiles} \cite{Google:Soli}, and apply then a peak search on this data.
The fast time processing involves a removal of DC components and a \ac{fft} along the distinct chirps (incl. dropping the symmetric part of the spectrum), which leads to complex range profiles.
Subsequently, the complex mean over the slow time axis is removed to get rid of (quasi-)static targets.
We integrate the magnitude of the complex data along the receive channels $R$ and chirps $C$ to improve the signal-to-noise ratio.
The resulting 1D vector of shape $\frac{S}{2} = 32$ represents the reflected energy of moving targets along the resolvable range.

\begin{figure}[!htbp]
    \centering
    \includegraphics[width=\columnwidth]{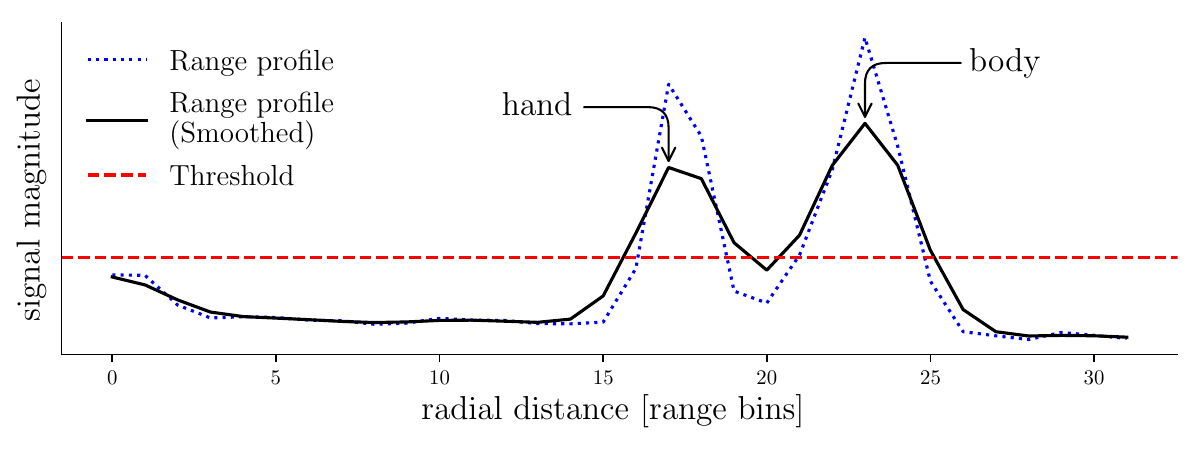}
    \caption{
        The range profile shows two targets above the threshold, the moving hand and the body of the person.
        Instead of selecting the target with the highest signal strength, the body, we select the closest target, the hand.
        This results in an efficient but stable target detection, which is immune to random body movements of the person performing the gesture.
    }
    \label{fig:target_detection}
\end{figure}

Based on the premise that the person performs the gesture towards the sensor, we can assume that the subject's hand is the closest moving target.
In such a setting, random body movements of the person are more distant than the hand, as shown in \cref{fig:target_detection}.
Therefore, we can run a local peak search and select the closest target, which returns its \textit{radial distance}.
We apply a Gaussian smoothing to the 1D vector and threshold the signal to suppress weak local maxima.
If all elements of the vector are below the threshold, e.g. if no person is in the field of view, we take the global maximum of the vector to generate a valid data point.

\subsection{Feature extraction}\label{chap:feature}
\begin{figure}[!htbp]
    \centering
    \includegraphics[width=\columnwidth]{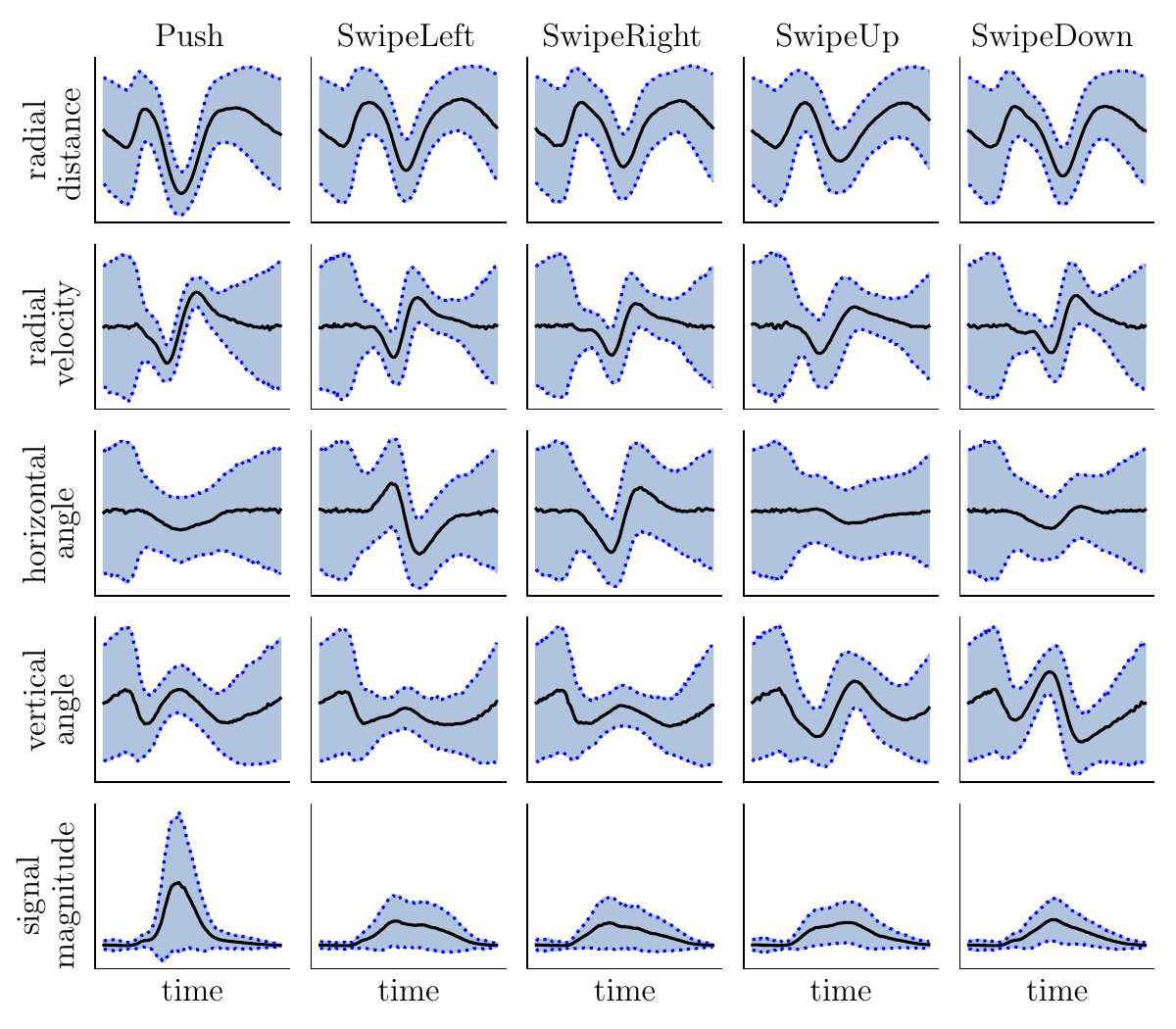}
    \caption{
        The matrix shows prototypical characteristics for the five gestures.
        The bold lines were generated by averaging 3200 samples per gesture; the shaded area around indicates a single standard deviation.
        The gestures differ in various characteristics, e.g. SwipeLeft and SwipeRight have contrarian series in the horizontal angle.
    }
    \label{fig:features}
\end{figure}

With the knowledge of the targets' range bin, we can extract the relevant scattering characteristics very efficiently.
Instead of performing slow time \acp{fft} across all range bins to generate \acp{rdi}, as in \cite{Google:Soli, Choi:LSTM, Google:RadarNet}, we perform only \acp{fft} on the detected range bin across all receive channels.
We integrate the resulting \textit{Doppler profile} across the channels and search for the maximum signal; its position indicates the \textit{radial velocity}, whereas the amplitude is the \textit{signal magnitude}.
Finally, we estimate the \textit{horizontal} and \textit{vertical angle} using phase-comparison monopulse on the detected Doppler bin \cite{Skolnik:Radar}.
\cref{fig:features} shows an overview of extracted features for each of the five gestures.

\section{Detection and Classification}
\subsection{Model architecture}

We detect and classify gestures based on a compact \ac{rnn} architecture.
The inputs for the network are at each timestamp the five extracted characteristics that we introduced in \cref{chap:feature}.
The network consists of only two layers - a \ac{gru} layer to capture temporal context, and a single dense layer with softmax activation for the classification.
The output layer contains six neurons, five for the gestures and one additional neuron to predict background.
We instantiate the network with 16 \acp{gru} (1104 parameters), followed by the dense layer with six neurons (102 parameters), which leads to a total number of 1206 parameters.

\subsection{Label refinement}
We found that a consistent labeling of gestures is essential for the performance of the overall system.
In addition to the correct label, this involves also the precise location inside the sequence.
During the creation of our dataset, the individuals were asked to perform a specific gesture every three seconds.
Hence, we can extract windows of 100 frames (approx. three seconds at a frame rate of $\SI{33.3}{\hertz}$) with a single, known gesture.
Afterwards, we dynamically refine the location of the label inside the extracted sequence.

\begin{figure}[!htbp]
    \centering
    \subfloat[Gesture samples]{\includegraphics[width=0.64\columnwidth]{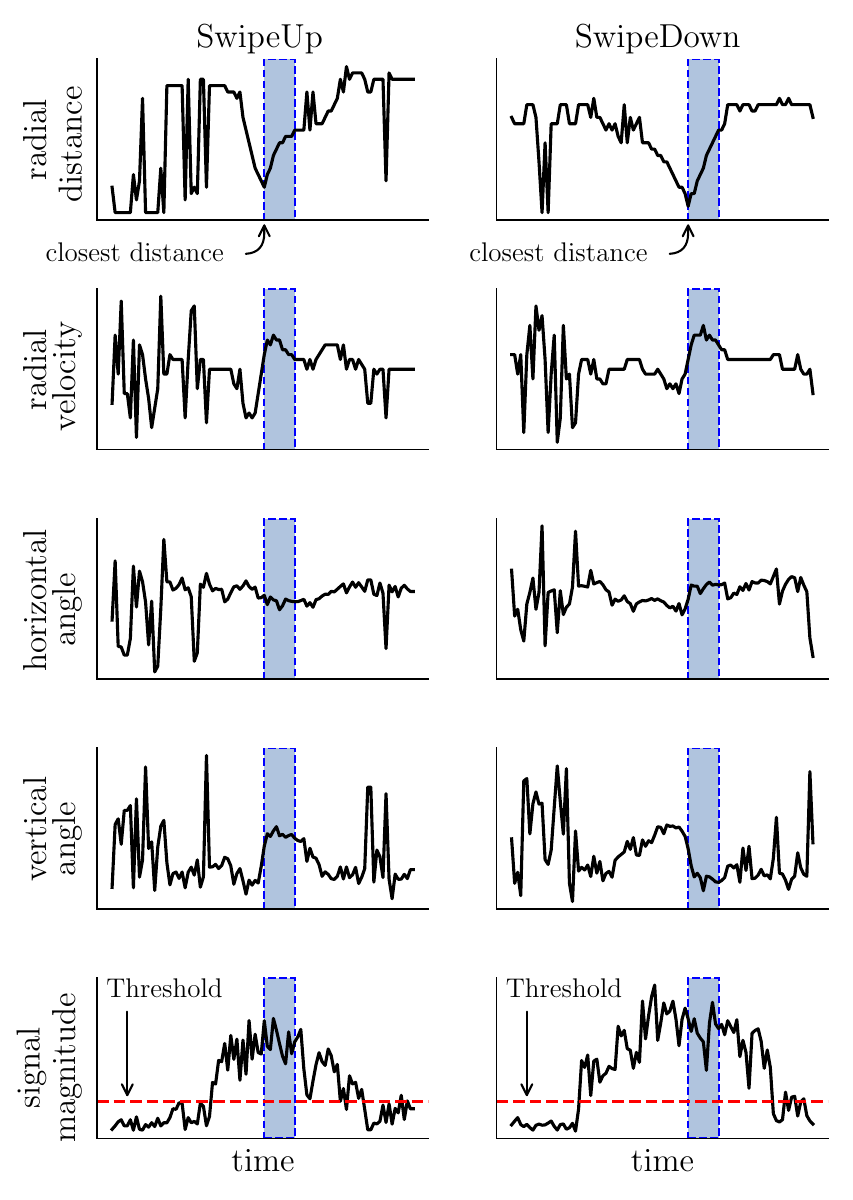}\label{fig:sample}}\hskip1ex
    \hfill
    \subfloat[Gesture sequence]{\includegraphics[width=0.34\columnwidth]{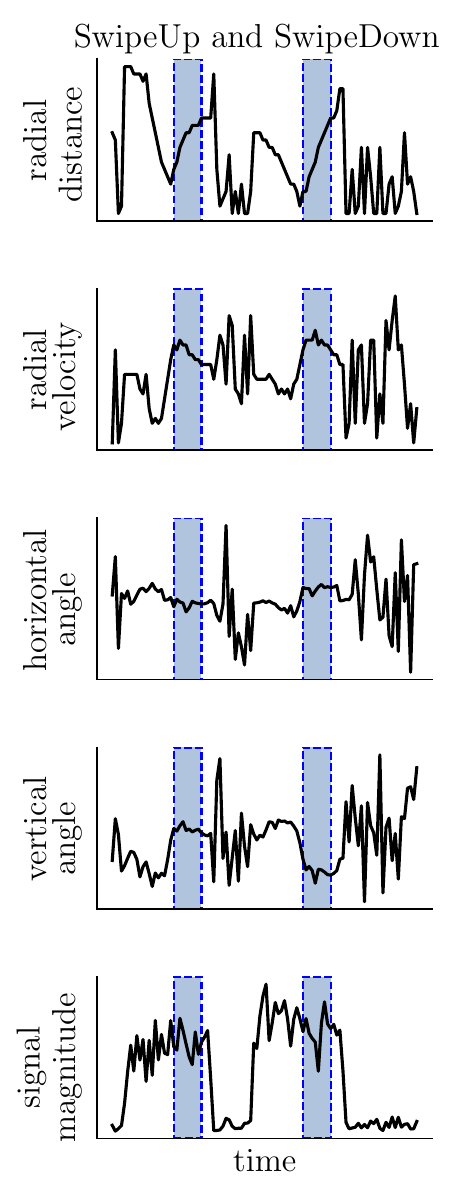}\label{fig:sequence}}
    \caption{
        \cref{fig:sample} shows the label refinement, where the closest distance above a specific amplitude threshold is shown.
        The emphasized area indicates the position of the gesture label; all other frames are labeled as background.
        The gesture samples from \cref{fig:sample} are used to compose an artificial gesture sequence, which is shown in \cref{fig:sequence}.
    }\label{fig:label}
\end{figure}

Our label refinement algorithm is based on the feature extraction described earlier in \cref{chap:feature}.
The supported gestures have in common that during their execution the hand reaches at one point a closest distance to the radar (see \cref{fig:features} and \cref{fig:sample}).
We employ this to fix the beginning of the gesture label at the frame where the radial distance is minimum.
In order to suppress false alarms at the beginning and end of sequences, where the noise dominates, we search for the minimum radial distance only at frames that have a signal amplitude above an empirically determined threshold.
The label is kept for a defined number of frames, before it changes to the background class, as depicted in \cref{fig:label}.

\subsection{Data augmentation}\label{sec:augmentation}
In order to increase the variety of the collected dataset, we apply a data augmentation strategy inspired by the audio domain \cite{Raju:Audio}.
We mix raw radar data from background sequences together with randomly sampled gestures - this allows us to create complex sequences of gestures, which would be difficult to record and label, compare \cref{fig:sequence}.
First, we use a background sequence that does not contain any gestures and sample multiple gestures, which will be inserted.
Afterwards, we inject the gestures randomly into the background sequence; it is ensured that gestures are not overlapping each other.
A smooth transition between background and gesture is established by weighting the raw data with a Tukey window.

\section{Experiments and Results}
\subsection{Dataset}\label{sec:dataset}
The dataset of this study was recorded with a BGT60TR13C XENSIV™ 60GHz radar sensor.
It includes gestures performed at different distances from the radar ($\SI{0.6}{\meter}$, $\SI{0.8}{\meter}$, $\SI{1.0}{\meter}$), three angles relative to the sensor ($-\SI{30}{\degree}$, $\SI{0}{\degree}$, $+\SI{30}{\degree}$), and a varying mounting height of the radar ($\SI{0.95}{\meter}$ - $\SI{1.35}{\meter}$).
The gestures were performed within a three second time window resulting in samples of 100 frames.
In addition, we recorded background sequences in which persons performed random body movements in front of the sensor to improve false alarm suppression.
The dataset contains in total 16k gestures and 3.4k background samples, which are used for training and validation, and nine sequences of 100 randomly sequenced gestures as test set.

\subsection{Gesture recognition performance}

We train the neural network with the dataset described in \cref{sec:dataset}.
The data is augmented with the strategy discussed in \cref{sec:augmentation} and split in a training and a validation set in a ratio of 3:1.
The network weights are fitted based on sequences of 100 time steps, i.e. input vectors of size $[100 \times 5]$, and label vectors of the same length, as illustrated in \cref{fig:label}.
We compute the categorical cross-entropy between the labels and the network predictions at each time step and accumulate it in the overall loss.
The model parameters are optimized with Adam, a learning rate of $\num{1e-3}$ and a batch size of 32.

\begin{figure}[!htbp]
    \centering
    \includegraphics[width=\columnwidth]{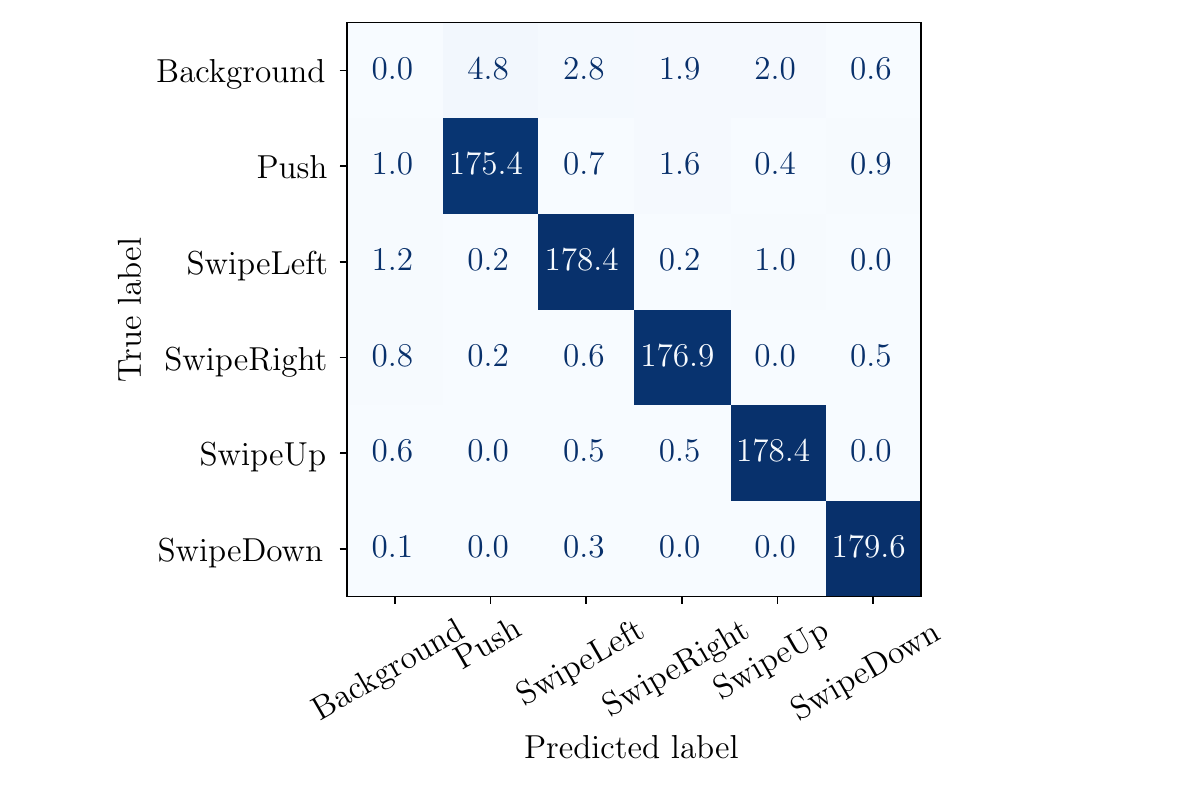}
    \caption{
        The confusion matrix shows the averaged results of ten randomly initialized training runs.
        Gestures mispredicted as background are false negatives, whereas background predicted as gesture is a false positive.
    }
    \label{fig:confusion}
\end{figure}

We run the training for 100 epochs and restore the model with the best validation accuracy.
The algorithm is evaluated on the hold-out test set.
Only if the algorithm predicts the correct gesture in a window of $\SI{150}{\milli\second}$ before and $\SI{300}{\milli\second}$ after the reference label, it is counted as correct.
The results of our evaluation are shown in \cref{fig:confusion} and lead to an F1 score of 98.4\%.
The low number of false positives and false negatives underlines the relevance of the system for practical applications.

\subsection{Gesture recognition on the edge}
The described system is implemented on a 32-bit PSoC™ 6 Arm® Cortex®-M4 to show that this approach is well suited for gesture sensing on the edge.
The feature extraction is based on the CMSIS-DSP software library, while we run the neural network inference with the TensorFlow Lite Micro interpreter.
Both the feature extraction and the neural network inference are computed in 32-bit single-precision floating-point numbers.
The whole system incl. data acquisition and application level logic uses a total of $\SI{120}{\kilo\byte}$ RAM, $\SI{278}{\kilo\byte}$ flash memory and CPU time of 33\%.
We measured a power consumption of $\SI{75}{\milli\watt}$ for the whole system.

\section{Conclusion}
In this work, we proposed a system to recognize hand gestures with a small computational footprint that fits on embedded processors.
A lightweight feature extraction algorithm combined with a slim neural network design enables robust gesture recognition for real-world applications.
Additionally, the label refinement and data augmentation harden the whole system to cope with the intended use cases.
We prove the robustness by evaluating the system on our hold-out test dataset.
Our future work includes investigations to extend the range of the system to higher distances and to enable more different gestures, but also to lower the resource consumption even more.


\bibliographystyle{IEEEtran}

\bibliography{IEEEabrv,IEEEexample}

\end{document}